\documentclass[11pt]{article}

\usepackage[preprint]{acl}

\usepackage{times}
\usepackage{latexsym}

\usepackage[T1]{fontenc}

\usepackage[utf8]{inputenc}

\usepackage{microtype}

\usepackage{inconsolata}

\usepackage{graphicx}

\usepackage{amsmath}
\usepackage{siunitx}
\usepackage{booktabs}
\usepackage{multirow}
\usepackage[table]{xcolor}
\title{SSR: Speculative Parallel Scaling Reasoning in Test-time}

\author{
Yuanlin CHU$^1$ $\qquad$ Bo WANG$^1$ $\qquad$ Xiang LIU$^1$ $\qquad$ Hong CHEN$^1$ \\ $\qquad$ \textbf{Aiwei LIU}$^2$ $\qquad$ \textbf{Xuming HU}$^{1,}$$^\dagger$\\
$^1$The Hong Kong University of Science and Technology (Guangzhou)\\
$^2$Tsinghua University\\
\texttt{\{ychu763,bwang423,xliu886,hchen763\}@connect.hkust-gz.edu.cn} \\
\texttt{liuaw20@mails.tsinghua.edu.cn, xuminghu@hkust-gz.edu.cn}
}

\begin{document}
\maketitle
\begin{abstract}
Large language models (LLMs) have achieved impressive results on multi-step mathematical reasoning, yet at the cost of high computational overhead. This challenge is particularly acute for test-time scaling methods such as parallel decoding, which increase answer diversity but scale poorly in efficiency. To address this efficiency--accuracy trade-off, we propose \textbf{SSR} (Speculative Parallel Scaling Reasoning), a training-free framework that leverages a key insight: by introducing speculative decoding at the \emph{step level}, we can accelerate reasoning without sacrificing correctness. SSR integrates two components: a \emph{Selective Parallel Module} (SPM) that identifies a small set of promising reasoning strategies via model-internal scoring, and \emph{Step-level Speculative Decoding} (SSD), which enables efficient draft--target collaboration for fine-grained reasoning acceleration. Experiments on three mathematical benchmarks—AIME 2024, MATH-500, and LiveMathBench—demonstrate that SSR achieves strong gains over baselines. For instance, on LiveMathBench, SSR improves pass@1 accuracy by 13.84\% while reducing computation to 80.5\% of the baseline FLOPs. On MATH-500, SSR reduces compute to just 30\% with no loss in accuracy.
\end{abstract}

\section{Introduction}

Large language models (LLMs) have shown strong capabilities in solving complex multi-step mathematical problems~\citep{wei2025surveyfeedbackbasedmultistepreasoning, wei2022chain, lewkowycz2022solving, cobbe2021trainingverifierssolvemath}, which require symbolic manipulation, structured planning, and precise stepwise reasoning~\citep{hendrycks2021measuringmathematicalproblemsolving}. A commonly adopted approach to improve accuracy is test-time scaling, which explores multiple reasoning trajectories via parallel decoding or tree search~\citep{ding2025dynamic, yu2025accelerateparallelizablereasoningparallel, wang2025harnessing}. However, these methods incur substantial computational overhead, producing an inherent efficiency–accuracy trade-off: improved reasoning quality often requires significantly higher compute, while aggressive efficiency control risks discarding promising reasoning paths.

This trade-off limits practical deployment in cost-sensitive settings, where indiscriminate path expansion frequently leads to redundant or low-quality computation~\citep{ding2025dynamic, yu2025accelerateparallelizablereasoningparallel}. Speculative decoding~\citep{leviathan2023fast, chen2023acceleratinglargelanguagemodel, sun2024block} offers efficiency improvements by letting a lightweight draft model propose outputs verified by a stronger target model, but its token-level verification primarily benefits short-form generation~\citep{chen2023revisiting} and is less suited for multi-step reasoning, where correctness depends on coherent step-level semantics~\citep{pan2025specreasonfastaccurateinferencetime, wang2025acceleratinglargelanguagemodel}.

Our key insight is to apply speculative decoding at the \emph{step level}, enabling the model to verify or revise complete reasoning chunks rather than individual tokens. Combined with selective parallel planning, this allows us to reduce redundant computation while preserving solution diversity. We therefore propose \textbf{SSR}, a training-free inference framework that addresses both sides of the efficiency–accuracy trade-off.

SSR consists of two components: a \textbf{Selective Parallel Module (SPM)} that selects a compact subset ($n \ll K$) of reasoning strategies from a curated pool, and \textbf{Step-level Speculative Decoding (SSD)}, where a draft model proposes reasoning steps that are verified and optionally revised by a target model. Experiments on multiple mathematical benchmarks demonstrate that SSR consistently improves efficiency–accuracy trade-offs across tasks, without model finetuning or additional training data.

\noindent\textbf{Contributions.}
We make the following contributions:
(1) we study the efficiency–accuracy trade-off in test-time reasoning;
(2) we propose SSR, integrating selective parallel scaling with step-level speculative reasoning in a training-free framework;
(3) we demonstrate consistent accuracy gains with significantly reduced computational cost across multiple benchmarks.

\section{Preliminary}
\subsection{Task Formulation}
We study \emph{multi-step mathematical reasoning}, where the goal is to generate a sequence of logically valid steps leading to a correct final answer. Formally, given a problem $x \in \mathcal{X}$, the model produces a sequence of reasoning steps $S = \{s_1, s_2, \dots, s_T\}$, followed by an answer $a \in \mathcal{A}$:
\begin{equation}
    S, a = f(x),
\end{equation}
where each $s_t$ represents a complete intermediate step, such as an algebraic simplification or numeric derivation. The final prediction is considered correct if $a$ matches the gold solution $\hat{a}$. Although intermediate steps are not graded, they serve to enhance interpretability and enable structured verification \cite{cobbe2021trainingverifierssolvemath, lewkowycz2022solving}.

Unlike direct answer generation, multi-step reasoning poses greater challenges for inference-time efficiency due to the extended sequence length and semantic granularity of each output unit.

\subsection{Scaling and Acceleration in LLM Reasoning}

Recent progress in LLM reasoning has followed two distinct directions:
\textbf{(1) Test-Time Scaling.} A widely used approach to improve accuracy is parallel decoding—sampling multiple reasoning paths simultaneously using different prompts, sampling seeds, or decoding strategies. While effective at increasing solution diversity, this strategy scales compute linearly with the number of paths, limiting its practical efficiency in multi-step settings.
\textbf{(2) Speculative Acceleration.} To improve inference efficiency, speculative decoding combines a lightweight draft model with a larger target model that verifies the generated outputs \cite{leviathan2023fast, chen2023acceleratinglargelanguagemodel}. Recently, emerging efforts have extended this idea to reasoning tasks, forming the basis of \emph{speculative reasoning}—leveraging draft–target collaboration to speed up multi-step inference. However, most prior work focuses on coarse-grained step verification and lacks fine-grained control over reasoning quality.

\begin{figure*}[ht]
    \centering
    \includegraphics[width=\textwidth]{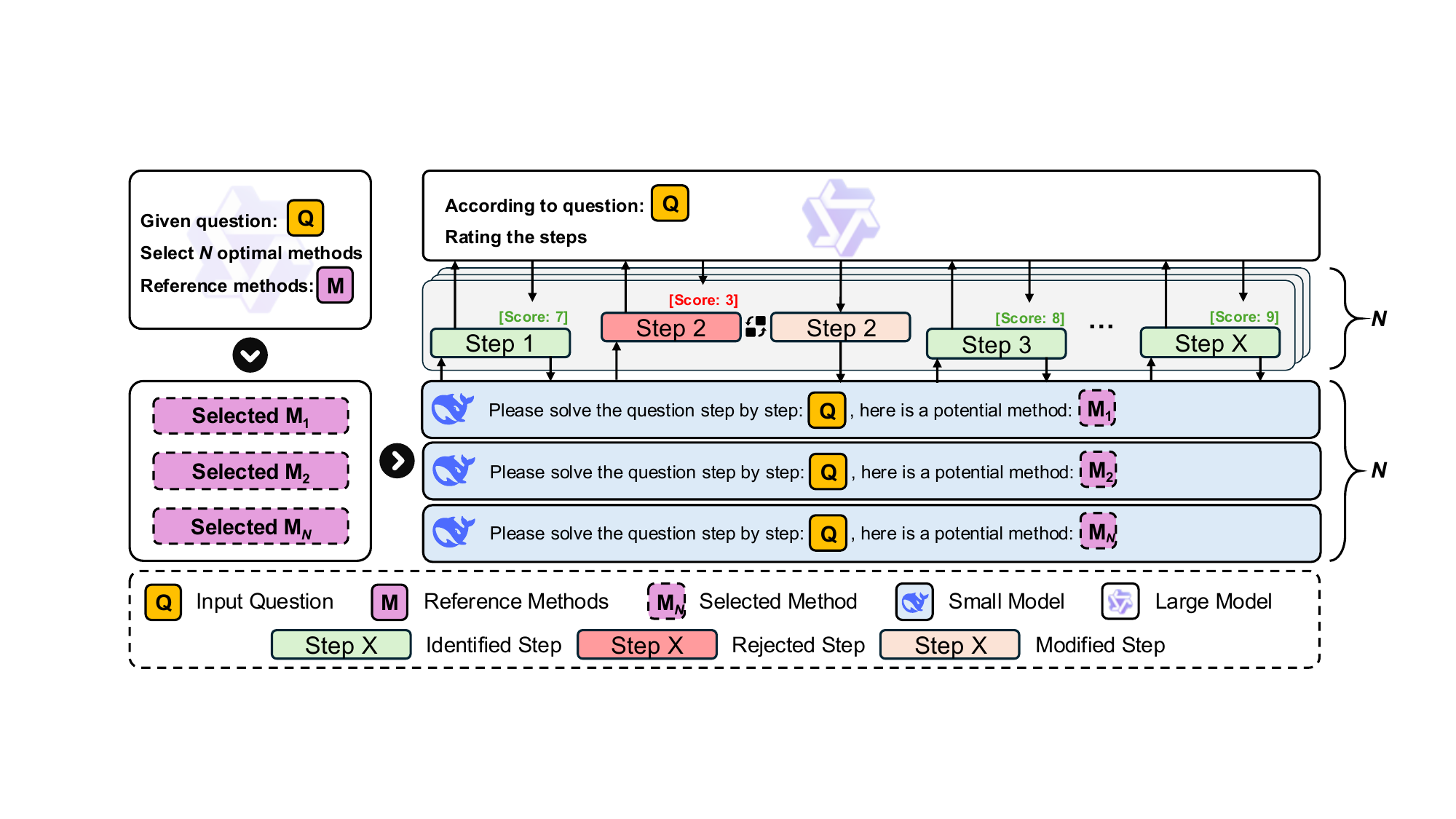}
    \caption{
        Overview of SSR framework. At test time, the target model first selects a subset of strategies from a curated strategy pool. For each selected method prompt, the draft model generates step-by-step reasoning. Each step is validated and optionally revised by the target model under a step-level speculative decoding scheme. All inference is executed in parallel with batched processing for efficiency.
    }
    \label{fig:pipeline}
\end{figure*}

\section{Methodology}

We propose a two-stage framework SSR that improves reasoning efficiency while maintaining solution quality: (1) \textbf{Selective Parallel Module} reduces the number of unnecessary reasoning paths by selecting promising strategies at test time; (2) \textbf{Step-level Speculative Decoding} accelerates the generation of each reasoning path by using a draft model to generate reasoning steps and a target model to verify them. Together, these techniques achieve substantial savings in inference cost.

\subsection{Selective Parallel Module}

\begin{figure}[ht]
    \centering
    \includegraphics[width=0.9\linewidth]{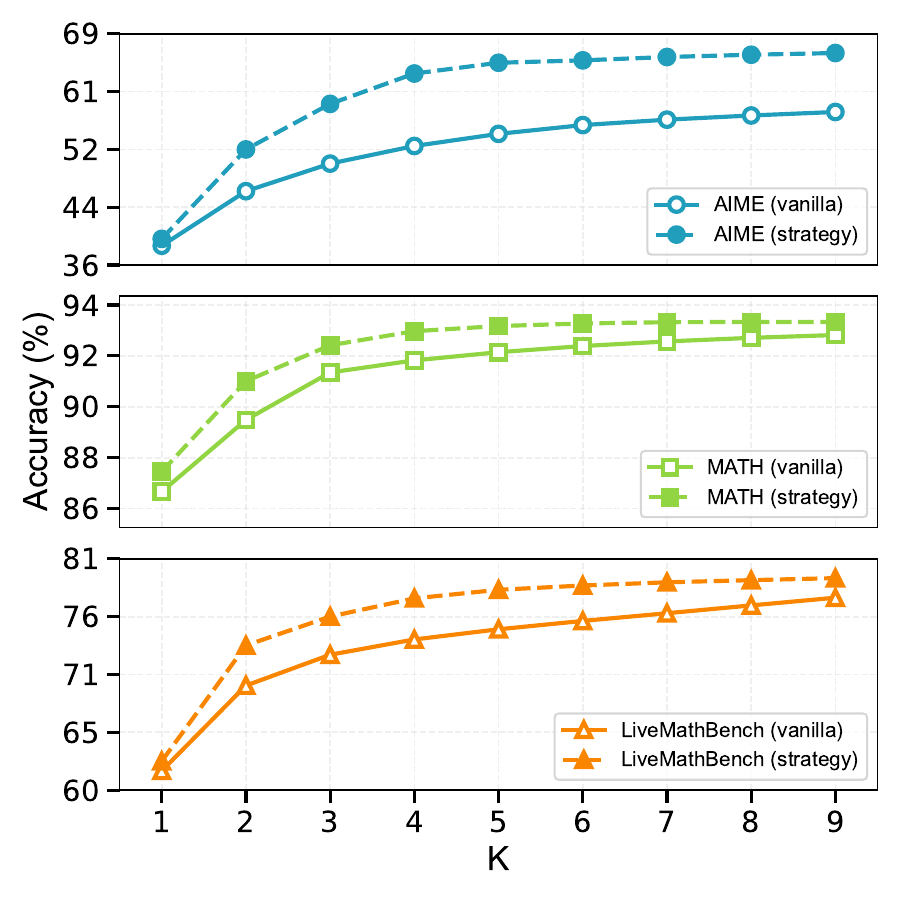}
    \caption{Accuracy (pass@1) vs. number of parallel reasoning paths on AIME, MATH-500, and LiveMathBench using Qwen/QwQ-32B. Each subplot compares two settings: (1) vanilla parallel decoding, and (2) strategy-aware decoding with diverse reasoning strategies.}
    \label{fig:scaling-accuracy}
\end{figure}

\noindent\textbf{Motivation.}
Parallel decoding improves accuracy by generating diverse reasoning paths, but its computational cost grows linearly with the number of paths. Figure~\ref{fig:scaling-accuracy} shows that although performance increases with more paths, it quickly saturates, meaning most additional paths incur redundant cost with limited gain. Moreover, when compared with naive parallel decoding, a strategy-aware variant that instantiates diverse reasoning strategies achieves comparable or superior accuracy using fewer paths, particularly under small $N$. These observations motivate SPM: instead of blindly expanding all candidates, we construct a curated strategy pool and selectively activate only a small, high-potential subset for each input.

\setlength{\belowcaptionskip}{-5pt}

\noindent\textbf{Strategy Pool Construction.}
We define a universal strategy pool $\mathcal{M} = \{m_1, \dots, m_K\}$, where each $m_i$ represents an interpretable math strategy (e.g., modular arithmetic, polynomial factorization, geometric construction). To ensure broad applicability, the pool is constructed to be task-agnostic: The same $K=12$ strategies are used across all datasets, consistently improving performance.

Importantly, the construction of $\mathcal{M}$ requires minimal manual effort. We first use a generative language model to summarize candidate strategies from model-generated solutions, followed by a lightweight verification stage by non-expert annotators. This low-cost process yields a transferable set of reasoning priors. We include further analysis in Appendix~C, showing that performance saturates as the pool size reaches 12. Additionally, Appendix~D provides concrete examples of the strategies used in our universal pool.

\noindent\textbf{Strategy Selection at Test Time.}
Rather than exhaustively executing all $K$ strategies per instance, we query the target model itself to select a subset of $n \ll K$ strategies most likely to yield correct reasoning. Specifically, we construct a multi-choice prompt listing all method names and their descriptions, and ask the model to return only $n$ identifiers it deems most suitable for the given problem. This near zero-cost control mechanism avoids unnecessary sampling and ensures that only high-quality paths are retained. 

\noindent\textbf{Parallel Execution.}
Once the $n$ strategies $\{m_{i_1}, \dots, m_{i_n}\}$ are selected, we run the reasoning model in parallel for each one, initializing the input as \texttt{[Problem Statement] + [Method Prompt]}. Since all reasoning paths now originate from semantically diverse and manually curated strategies, they exhibit high variability without random sampling. All decoding executions are batched to leverage GPU acceleration, and the overall compute is reduced proportionally from $K$ paths to $n$ selected ones.

\subsection{Step-level Speculative Decoding}

\noindent\textbf{Overview.}
To reduce the computational cost of large reasoning models, we propose a step-level speculative decoding framework in which a lightweight draft model generates full reasoning steps, and a stronger target model verifies and selectively rewrites them. Compared with traditional token-level speculative decoding, our design explicitly aligns the verification granularity with the semantic structure of mathematical reasoning: correctness is often determined at the level of complete logical steps rather than individual tokens. Operating on step units therefore avoids excessive micro-level validation, reduces redundant computation, and preserves global reasoning coherence. In addition, step-level operation naturally supports batching and efficient parallel execution, while remaining compatible with recent findings that only a sparse set of high-entropy signals dominates reasoning quality~\cite{wang2025beyond}.

\noindent\textbf{Step-wise Validation and Revision.}
Let $M_d$ and $M_t$ denote the draft and target models. For each reasoning path, $M_d$ generates a step $s_t$ conditioned on $\{s_1,\dots,s_{t-1}\}$. The target model assigns a plausibility score:
\begin{equation}
    \text{score}(s_t) = M_t(x, s_1, \dots, s_t)
\end{equation}
using a lightweight single-token response. With a threshold $\tau \in [0,9]$, steps with $\text{score}(s_t) \ge \tau$ are accepted; otherwise $M_t$ rewrites $s_t$ into $s'_t$, from which decoding resumes.

\noindent\textbf{Parallel Batched Inference.}
As shown in Figure~\ref{fig:pipeline}, all candidate reasoning paths are decoded in parallel. For each step index, we batch the forward pass of $M_d$, followed by batched scoring or rewriting using $M_t$, enabling efficient large-batch speculative evaluation.

\noindent\textbf{Answer Aggregation Strategy.}
After decoding, we aggregate outputs across paths. By default we apply Majority Voting. When answers disagree, we employ a score-based strategy inspired by PRM: paths are ranked by their mean step score, with rewritten steps assigned a confidence of $9$.

\noindent\textbf{Fast Modes.}
For latency-sensitive applications, we further introduce two early-exit variants that allow the system to terminate decoding before all speculative paths are completed: 
\textbf{Fast-1} stops once any path produces a final answer, providing the fastest response, while 
\textbf{Fast-2} terminates when two paths converge to the same answer, offering a stronger confidence–efficiency balance. 
These variants substantially reduce computation in practical deployment scenarios while retaining competitive accuracy.

\noindent\textbf{Theoretical Resource Estimation.}
Let $L$ denote the average step length, and $C_d$ and $C_t$ the per-step cost of the draft and target models. If a fraction $R$ of steps are rewritten, the per-step cost is
\begin{equation}
C_{\text{step}} = C_d + R \cdot C_t.
\end{equation}
Let $K$ be the full number of parallel paths and $n$ the selected subset. The normalized compute ratio is:
\begin{equation}
\gamma = \frac{n}{K}\left(\frac{C_d + R \cdot C_t}{C_t}\right),
\end{equation}
showing that combining selective parallelism with step-level speculation yields substantial efficiency gains when $n \ll K$ and the draft model is cheap.

\section{Experiments}

\subsection{Experimental Settings}

\noindent\textbf{Datasets.}
We evaluate our approach on three representative math reasoning benchmarks: AIME 2024, MATH-500, and LiveMathBench (AMC\_en subset), which collectively span olympiad-level symbolic problems, textbook-style multi-step derivations, and linguistically diverse real-world questions.

\noindent\textbf{Models.}
We use QwQ-32B as the target model ($M_t$) and DeepSeek-R1-Distill-Qwen-1.5B as the draft model ($M_d$). This choice is motivated by the following considerations:
\textbf{(1) Compute gap}: The two models differ substantially in parameter scale (32B vs.\ 1.5B), which enables clear potential for compute savings. 
\textbf{(2) Capability balance}: Despite the scale gap, both models achieve reasonable accuracy on math tasks, allowing the smaller model to generate plausible steps while the larger model retains strong revision capacity. 
\textbf{(3) Distribution alignment}: DeepSeek-R1-Distill-Qwen-1.5B is distilled from Qwen, ensuring alignment in token distributions, which is essential for speculative decoding to be effective.

\noindent\textbf{Parameter Settings.}
To ensure fair comparison and computational efficiency, we limit all generations to a maximum length of 8192 tokens. As our focus is on improving reasoning efficiency rather than modeling extremely long outputs, we exclude overly long sequences from evaluation. This constraint significantly reduces the available context window and output space for complex problems, especially on benchmarks like AIME2024, where full solutions often exceed 10,000 tokens. As a result, the absolute accuracy of the large target model (e.g., QwQ-32B) is notably lower than its reported performance in unconstrained settings~\cite{qwq32b}, but this ensures a controlled and realistic comparison across all decoding strategies.

For step-level speculative decoding (SSD), we set the rewrite threshold to 7. This value was chosen empirically based on the distribution of step scores across datasets. Specifically, analysis shows that steps with scores below 7 constitute just over 20\% of all cases in AIME, MATH, and LiveMathBench. By setting the threshold to 7, we rewrite only the bottom ~20\% of low-confidence steps, preserving the majority of correct reasoning while maintaining efficiency. This strikes a favorable balance between computational savings and semantic fidelity.

We further validate this choice via an ablation study on AIME, varying the threshold and measuring its effect on both accuracy and compute cost. Results, provided in Appendix~B, confirm that threshold~7 achieves the best trade-off.

\begin{figure}[ht]
    \centering
    \includegraphics[width=\linewidth]{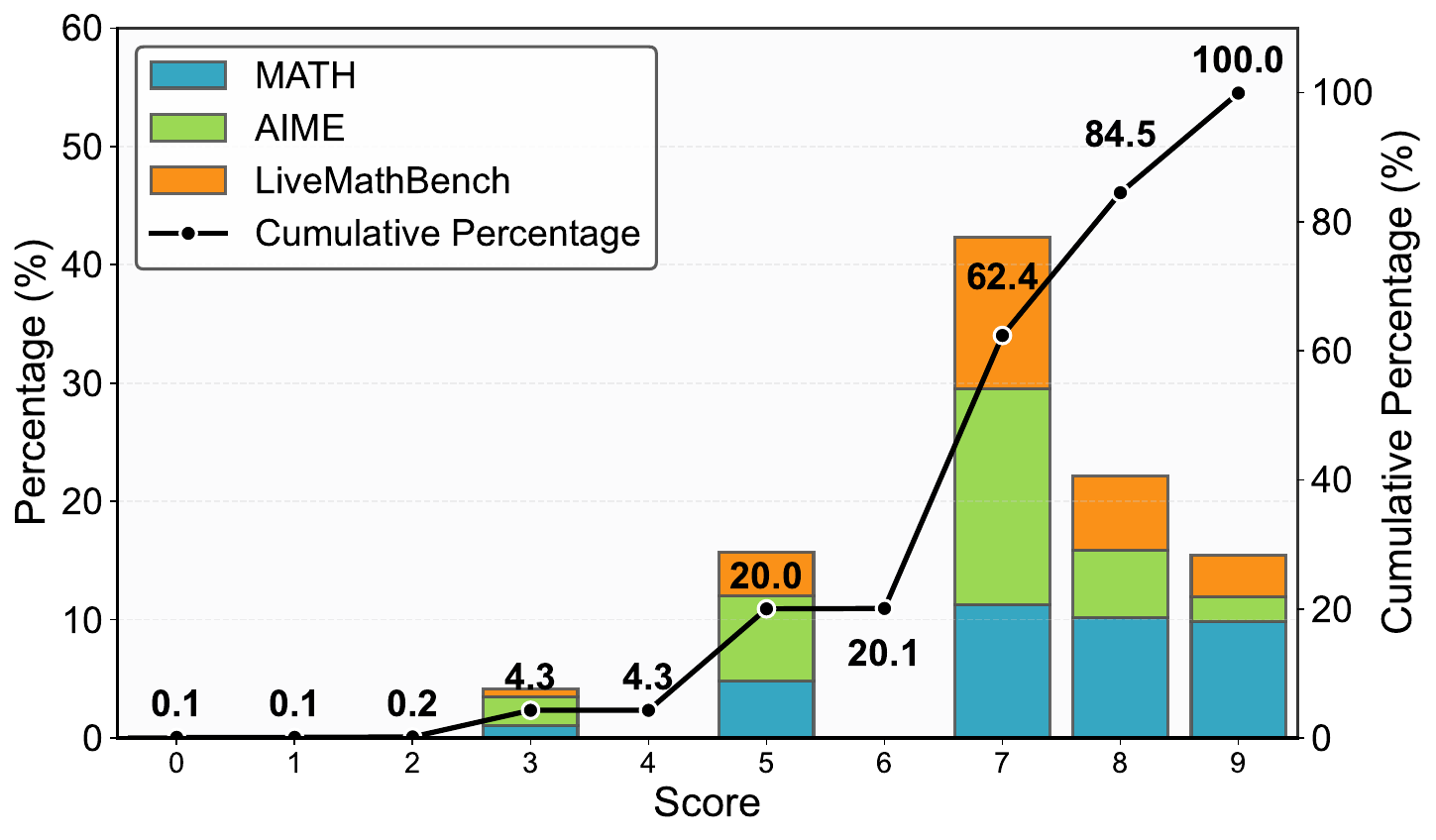}
    \caption{Distribution of step score (0–9) across AIME, MATH, and LiveMathBench using SSD. Bars show per-score proportions; the curve denotes cumulative percentage.}
    \label{fig:score_distribution}
\end{figure}

\noindent\textbf{Evaluation Metrics.}
We evaluate each inference method along three axes:

\textbf{(1) Accuracy}. We report \textit{pass@$k$} accuracy, defined as the fraction of problems for which the correct answer appears in the top $k$ generated candidates. Specifically, pass@1 corresponds to standard exact-match accuracy (i.e., whether the top-1 answer matches the ground truth), while larger $k$ values reflect the model's ability to include the correct answer among multiple sampled outputs. In this work, we primarily report pass@1 for comparability with prior work, and use pass@3 as auxiliary metrics to assess the diversity and reliability of speculative decoding.

\textbf{(2) Latency}. We report average inference latency per query measured on a 4×A800 (80GB) cluster.

\textbf{(3) Normalized FLOPs}.
Traditional efficiency metrics such as latency or throughput are hardware-dependent and implementation-sensitive. To enable hardware-agnostic comparison, we adopt a compute-efficiency metric called \textit{Normalized FLOPs} ($\gamma$), which estimates relative inference cost across strategies.

We define $\gamma = 1$ for standard single-path inference. For traditional multi-path inference with $N$ independent decoding branches, the compute cost scales linearly: $\gamma_{\text{parallel}} = N$.

For SSR-style speculative inference with $N$ paths, draft-to-target cost ratio $\alpha$, and rewrite rate $R$, the relative cost is:

\begin{equation}
\boxed{
\gamma_{\text{ssr}} = N\,\beta\,(R + (1 - R)\,\alpha)
}
\end{equation}

where $\beta = T/T_{\text{base}}$ denotes the relative number of tokens generated per path. This closed-form allows direct, model-agnostic comparison of speculative versus traditional decoding efficiency. A detailed derivation of $\gamma$ is provided in Appendix~A.

\begin{figure*}[t]
    \centering
    \includegraphics[width=\textwidth]{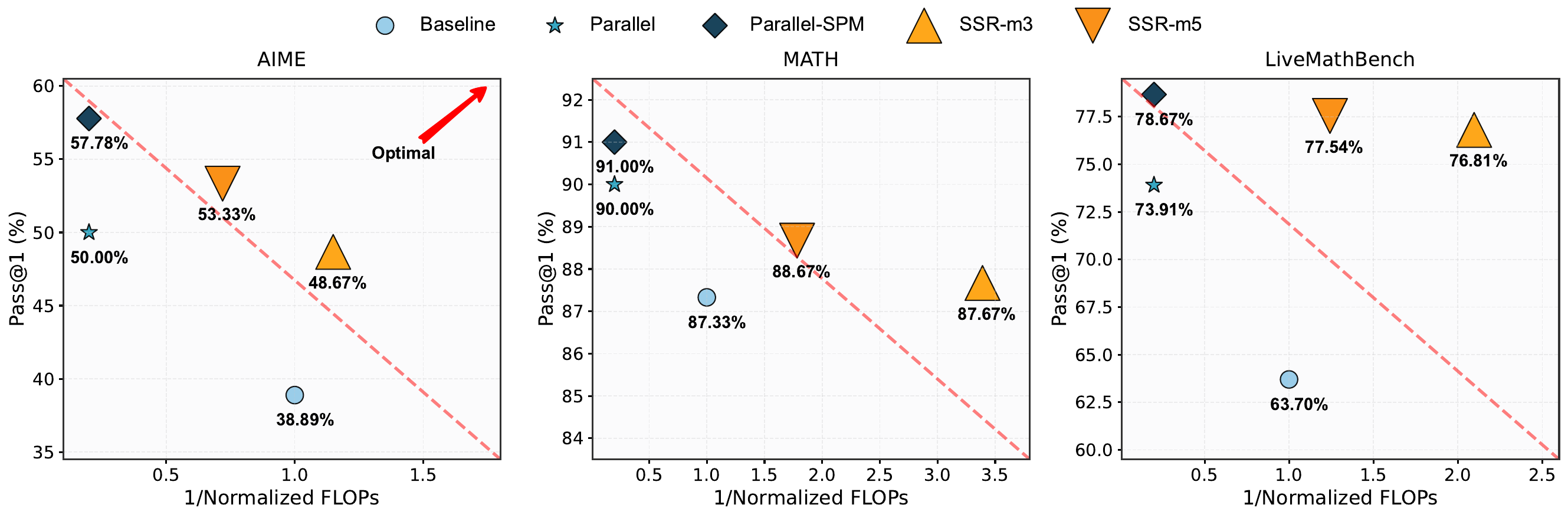}
    \caption{
    Comparison of different inference strategies on AIME2024, MATH-500, and LiveMathBench. Each sub-plot shows the trade-off between Computational Efficiency (x-axis: inverse of normalized FLOPs; higher is better) and Accuracy (y-axis: pass@1; higher is better). The red dashed line represents the Pareto frontier—only methods above this line are considered Pareto-optimal, offering better accuracy for a given compute budget. Points closer to the top-right corner indicate more desirable performance in terms of both accuracy and efficiency.
    }
    \label{fig:overall_results}
\end{figure*}

\subsection{Overall Results}

To assess the overall performance of our method, we compare five decoding strategies across three benchmark datasets: AIME2024, MATH-500, and LiveMathBench (AMC\_en). The evaluated settings include:

\begin{itemize}
    \item \textbf{Baseline}: Standard single-path decoding using the large model.
    \item \textbf{Parallel}: Naive parallel decoding without prompts, with $N=5$ paths.
    \item \textbf{Parallel-SPM}: Parallel decoding with SPM, with $N=5$ paths.
    \item \textbf{SSR-m3}: SSR framework with 3 speculative parallel paths.
    \item \textbf{SSR-m5}: SSR framework with 5 speculative parallel paths.
\end{itemize}

\noindent
In Figure~\ref{fig:overall_results}, we compare different inference strategies in terms of pass@1 accuracy versus normalized computational cost. All reported metrics are averaged over 6 runs.

The \textbf{Parallel} method, especially when combined with our SPM, delivers the highest accuracy across all benchmarks. For example, on MATH and LiveMathBench, \textbf{Parallel-SPM} achieves 91.00\% and 78.67\% pass@1, outperforming all other methods. However, this comes at a steep cost: up to 5$\times$ more FLOPs compared to standard decoding. While this approach improves correctness by exhaustively exploring multiple reasoning paths, the significant compute overhead makes it less practical for resource-constrained deployment.

In contrast, our proposed SSR framework, particularly \textbf{SSR-m3} and \textbf{SSR-m5}, achieves a favorable trade-off between accuracy and efficiency. On MATH, SSR-m3 reduces compute to only 30\% of the baseline while slightly improving accuracy. On LiveMathBench, SSR-m3 improves accuracy by +13.11\% with only 48\% of the original FLOPs, and SSR-m5 achieves +13.84\% accuracy gain using just 80.5\% of baseline cost. Although SSR yields slightly lower accuracy than Parallel-SPM, its compute savings are substantial—often exceeding 4--5$\times$—making it highly cost-effective.

On the more challenging AIME2024 dataset, where problem complexity likely exceeds the capability of the draft model, the benefit of speculative decoding is less pronounced. Nevertheless, SSR-m3 and SSR-m5 still improve accuracy by +9.78\% and +14.44\%, respectively, over baseline decoding, with only moderate increases in compute.

These results demonstrate that while aggressive parallel decoding can maximize accuracy, SSR methods offer a compelling middle ground: strong performance with drastically reduced compute. By combining speculative decoding with selective rewriting and multi-path scaling, SSR effectively mitigates the limitations of imperfect draft models without incurring the full cost of parallel inference.

\noindent\textbf{Summary.}
Our SSR framework demonstrates a clear advantage in efficiency--accuracy trade-off. While speculative decoding introduces slight degradation in raw accuracy compared to naive parallel inference, it drastically reduces computational cost. With method-level filtering and confidence-aware rewriting, SSR is able to outperform the baseline both in accuracy and efficiency on almost all tested benchmarks.

\subsection{Effectiveness of SPM}

To further validate the contribution of our SPM strategy, we conduct a focused ablation study comparing pure parallel inference against its SPM-augmented counterpart. In both settings, we fix the number of inference paths to $N=5$ and disable the SSD strategy to isolate the effect of SPM alone. Metrics are averaged over 6 independent runs.

\begin{figure}[ht]
    \centering
    \includegraphics[width=0.9\linewidth]{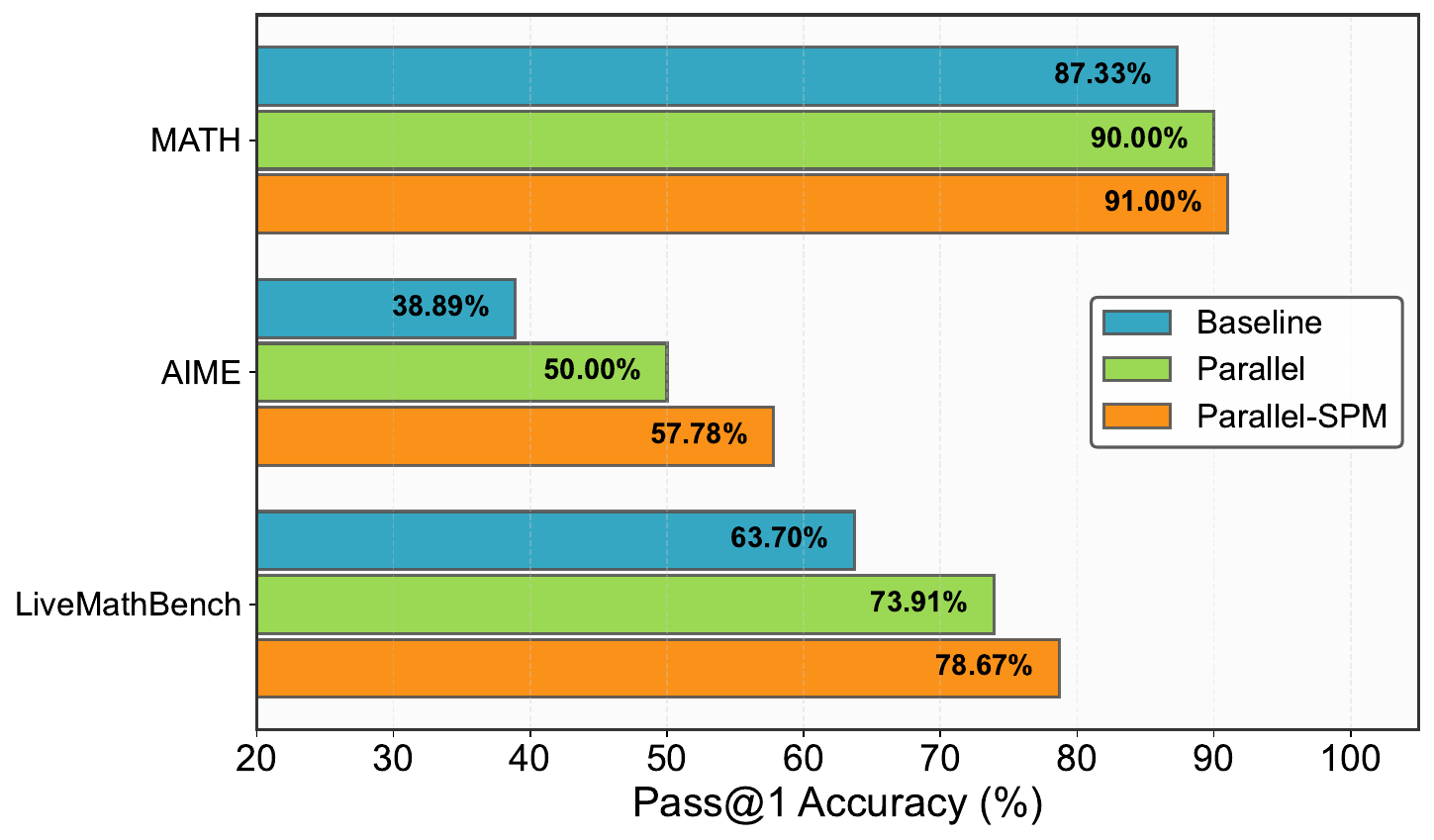}
    \caption{Ablation study on the impact of the Selective Parallel Module (SPM) across datasets. Each group of bars compares the Baseline, Parallel, and Parallel-SPM settings (without SSD; $N{=}5$ paths).}
    \label{fig:spm_ablation}
\end{figure}

\sisetup{
    detect-all,
    round-mode=places,
    round-precision=2,
    table-format=2.2,
    separate-uncertainty
}

\begin{table*}[ht]
\centering
\small
\begin{tabular}{
    l
    S[table-format=2.2] S[table-format=2.2] S[table-format=3.2]
    S[table-format=2.2] S[table-format=2.2] S[table-format=2.2]
    S[table-format=2.2] S[table-format=2.2] S[table-format=3.2]
}
\toprule
\multirow{2}{*}{\textbf{Method}} &
\multicolumn{3}{c}{\textbf{AIME2024}} &
\multicolumn{3}{c}{\textbf{MATH-500}} &
\multicolumn{3}{c}{\textbf{LiveMathBench}} \\
& {pass@1$\uparrow$} & {pass@3$\uparrow$} & {Time$\downarrow$}
& {pass@1$\uparrow$} & {pass@3$\uparrow$} & {Time$\downarrow$}
& {pass@1$\uparrow$} & {pass@3$\uparrow$} & {Time$\downarrow$} \\
\midrule
\rowcolor{gray!10} baseline         & 38.89 & 50.00 & 123.61 & 87.33 & 92.00 & 69.82 & 63.70 & 73.00 & 107.53 \\
spec-reason (7)                    & 32.22 & 53.33 &  83.16 & 76.00 & 88.00 & 19.47 & 60.87 & 78.26 &  50.15 \\
spec-reason (9)                    & 47.78 & 56.67 & 147.48 & 78.00 & 90.00 & 44.33 & 70.29 & 80.44 & 106.94 \\
SSR-Fast-1                         & 45.56 & \textbf{63.33} & 201.08 & 87.78 & 91.00 & 36.82 & 68.12 & 76.09 & 102.00 \\
SSR-Fast-2                         & 50.00 & \textbf{63.33} & 229.59 & \textbf{88.67} & \textbf{92.00} & 47.63 & 75.36 & 82.61 & 123.63 \\
\rowcolor{gray!15} SSR (Full)      & \textbf{53.33} & \textbf{63.33} & 259.00 & \textbf{88.67} & \textbf{92.00} & 67.07 & \textbf{77.54} & \textbf{84.78} & 151.65 \\
\bottomrule
\end{tabular}
\caption{Comparison between baseline, spec-reason, and SSR variants. All SSR methods use $N{=}5$ paths and threshold $7$. All reported metrics are averaged over 6 runs.}
\label{tab:ssr_vs_spec}
\end{table*}
\normalsize

As shown in Figure~\ref{fig:spm_ablation}, our SPM consistently improves accuracy over naïve parallel decoding. On AIME2024, SPM improves the pass@1 score from 50.00\% to 57.78\%; on MATH-500, accuracy increases from 90.00\% to 91.00\%; and on LiveMathBench, accuracy rises from 73.91\% to 78.67\%. Despite the already high baseline of the MATH dataset, SPM still yields measurable gains.

These results demonstrate that selectively activating decoding paths based on our scoring heuristics can significantly reduce error accumulation introduced by parallel speculative generation. The improvements are especially pronounced on more challenging datasets like AIME2024 and LiveMathBench, where the benefits of pruning less promising paths are more apparent.

\subsection{Comparing with Speculative Reasoning}

To further assess the effectiveness of our Speculative parallel Scaling Reasoning (SSR) framework, we compare it against a speculative reasoning baseline (\textbf{spec-reason}) inspired by prior work \cite{pan2025specreasonfastaccurateinferencetime}. While both approaches rely on a confidence threshold to determine whether to reuse or rewrite tokens, spec-reason performs \emph{sequential} speculative decoding with no multi-path coordination, and thus avoids the complexity of answer aggregation and path scheduling.

We evaluate spec-reason under two different thresholds (7 and 9), and compare it to three variants of our SSR framework: SSR-Fast-1 and SSR-Fast-2 refer to our two proposed fast modes (Fast Mode 1 and Fast Mode 2, respectively), and SSR denotes the full framework. All SSR variants use five parallel reasoning paths (i.e., $N=5$) and a rewrite threshold of 7. We report pass@1, pass@3, and average inference time (in seconds) on three datasets.

\noindent\textbf{Main Findings.}
From Table~\ref{tab:ssr_vs_spec}, we observe that our SSR framework outperforms both the baseline and speculative reasoning methods across all available datasets. In particular, SSR achieves the highest accuracy on all three datasets in both pass@1 and pass@3, demonstrating the benefit of combining speculative decoding with multi-path aggregation and dynamic scoring.

Although spec-reason can offer inference speedups, its accuracy suffers due to its lack of parallelism and coordination. For instance, with a threshold of 7, spec-reason yields only 32.22\% pass@1 on AIME2024, compared to 53.33\% from SSR—a 21.1\% absolute gain. Increasing the threshold (to 9) improves spec-reason's accuracy but at a substantial cost to latency, highlighting a less efficient trade-off than SSR.

\noindent\textbf{Fast Modes Effectiveness.}
We also find our proposed fast decoding modes, SSR-Fast-1 and SSR-Fast-2, to be highly effective. On relatively easier datasets such as MATH-500, SSR-Fast-1 achieves comparable accuracy 87.78\% to the full SSR framework 88.67\%, but at nearly half the inference time (36.82s vs. 67.07s). This suggests that our fast modes are well-suited for tasks where high throughput is desired without significant accuracy degradation.

\noindent\textbf{Adaptive Trade-offs.}
One key strength of the SSR framework is its ability to adaptively balance accuracy and efficiency based on input difficulty. On harder datasets such as AIME2024, SSR tends to invest more decoding effort to ensure higher precision. On easier benchmarks like MATH-500, it naturally achieves faster inference with minimal accuracy compromise. This is enabled by our dynamic scoring and rewriting strategy, which automatically scales computation according to token uncertainty and problem difficulty.

Importantly, SSR is not designed to endlessly boost model capability beyond its inherent capacity. Instead, it aims to help the model reach its performance ceiling more efficiently, with fewer wasted attempts. The ultimate performance is still bounded by the underlying model’s reasoning ability.

\section{Related Work}
Our work intersects two prominent lines of research in natural language processing: test-time scaling for reasoning models and speculative decoding techniques. We explore how speculative reasoning can be integrated into parallel scaling frameworks to enhance both efficiency and accuracy.

\subsection{Test-Time Scaling in Reasoning Models}

Recent work has shown that Test-Time Scaling (TTS) can substantially improve the reasoning abilities of large language models (LLMs) without any additional training. These strategies can be categorized into \textit{parallel scaling}, \textit{sequential scaling}, and their combinations.

Parallel scaling typically involves generating multiple reasoning paths in parallel and selecting the most consistent or likely answer. For example, the S* framework~\cite{li2025s} proposes a hybrid parallel-sequential decoding scheme for code generation, yielding strong improvements across several LLM backends. Sequential scaling, in contrast, encourages multi-stage reasoning through iterative prompting. The ``Think Twice'' method~\cite{tian2025think} prompts the model to reflect and revise its initial prediction, leading to consistent gains on complex math problems.

Hybrid approaches that combine these two axes have also been explored. The m1 framework~\cite{huang2025m1unleashpotentialtesttime} integrates parallel and sequential strategies in medical diagnosis tasks and achieves competitive performance even compared to much larger models. Liu et al.~\cite{liu2025can} demonstrate that a 1B-parameter model with appropriately scaled test-time inference can outperform models over 400$\times$ larger on challenging mathematical reasoning benchmarks. These findings underline the importance of TTS as a scalable and model-agnostic mechanism to enhance LLM reasoning.

\subsection{Speculative Reasoning}

Speculative decoding is an emerging paradigm for accelerating inference. Originally introduced by \citet{leviathan2023fast}, this method achieves 2--3$\times$ speedups on sequence generation tasks without sacrificing output quality. Subsequent works have refined this framework from both system and algorithmic perspectives. FastDecode~\cite{he2024fastdecodehighthroughputgpuefficientllm} designs a CPU–GPU hybrid pipeline to maximize speculative throughput under hardware constraints. Medusa~\cite{cai2024medusa} augments the decoder with parallel token heads and a tree-based attention mechanism, achieving up to 3.6$\times$ decoding acceleration. Sun et al.~\cite{sun2024block} propose block-wise verification to validate entire token chunks at once, yielding additional 5--8\% speedups.

More recently, speculative techniques have been extended from token-level prediction to multi-step reasoning. Wang et al.~\cite{wang2025efficient} introduce \textsc{SCoT}, where a small model generates complete chain-of-thoughts (CoTs), and the target LLM selects the most promising draft for final output. In a related line, SpecReason~\cite{pan2025specreasonfastaccurateinferencetime} allows the draft model to suggest intermediate reasoning steps, which are verified or corrected by the large model. Remarkably, it not only achieves 1.5--2.5$\times$ speedups but also improves answer accuracy by up to 9.9\%.

Overall, these approaches demonstrate that speculative decoding is not limited to accelerating surface-level token generation, but can be generalized to the reasoning process itself, enabling efficient yet accurate inference for complex tasks.

\section{Conclusion}

In this work, we present a novel integration of speculative reasoning into the test-time scaling paradigm, with the goal of alleviating the heavy computational burden typically associated with parallel decoding strategies. Our proposed framework consists of two complementary components: \textit{Selective Parallel Module} (SPM) and \textit{Step-level Speculative Decoding} (SSD).

SPM enhances reasoning efficiency by first selecting a promising reasoning strategy from a pre-constructed strategy pool, then dynamically constructing the corresponding prompt. This selection-before-prompting pipeline encourages both diversity and correctness among generated reasoning paths, without the need for additional training.

Building upon this, SSD applies speculative verification at the granularity of individual reasoning steps across multiple candidate paths. This fine-grained speculative decoding significantly reduces unnecessary computation by allowing early acceptance or rejection of intermediate steps. Furthermore, our scoring mechanism enables the system to adaptively trade off accuracy and efficiency within the SSR framework, while an additional \textit{fast mode} further accelerates inference for latency-sensitive applications.

Extensive experiments on multiple math reasoning benchmarks demonstrate that our method consistently improves both final-answer accuracy and decoding efficiency, despite being entirely training-free. These results highlight the potential of speculative reasoning as a powerful tool for efficient test-time computation in complex reasoning tasks.

\section*{Limitations}

Our approach currently focuses on mathematical reasoning tasks, which benefit from their structured and compositional nature. While this setting provides a clear foundation for testing speculative reasoning, generalization to domains with less rigid structure remains to be explored.
Additionally, the parallel speculative process introduces non-trivial scheduling and resource allocation challenges. Due to suboptimal utilization in our current implementation, the theoretical efficiency gains are not fully achieved. Future work will extend our framework to broader reasoning tasks and improve system-level optimizations for better runtime performance.

\bibliography{custom}

\appendix

\section{Normalized FLOPs Derivation}
\label{app:flops_analysis}

To fairly compare inference-time compute under different decoding regimes, we derive analytical expressions for the relative cost in floating-point operations (FLOPs), normalized against the baseline.

\subsection{Notation}
We summarize the key symbols used in our computational cost analysis as shown in Table~\ref{tab:description}.

\begin{table}[htbp]
    \centering
    \footnotesize
    \resizebox{0.48\textwidth}{!}{
    \begin{tabular}{ll}
    \toprule
    \textbf{Notion} & \textbf{Description} \\
    \midrule
        $N$ & Number of parallel inference paths \\ [0.3em]
        $T_{\mathrm{base}}$ & Tokens generated by the target model \\ [0.3em]
        $T$ & Tokens processed per path (speculative) \\ [0.3em]
        $\beta$ & Relative token count, defined as $\dfrac{T}{T_{\mathrm{base}}}$ \\ [0.3em]
        $F_{t}$ & FLOPs per token for the target model \\ [0.3em]
        $F_{d}$ & FLOPs per token for the draft model \\ [0.3em]
        $\alpha$ & Relative compute cost, defined as $\dfrac{F_d}{F_t}$ \\ [0.3em]
        $R$ & Rewrite rate (revised by $M_t$) \\ [0.3em]
    \bottomrule
    \end{tabular}
    }
    \caption{Description of Notions}
    \label{tab:description}
\end{table}

\subsection{FLOPs Ratio Estimation Per Token}

In this section, we estimate the theoretical FLOPs required to generate a single token by two transformer-based models: \textbf{DeepSeek-R1-Distill-Qwen-1.5B} and \textbf{QwQ-32B}. The estimation follows standard practice~\cite{kaplan2020scalinglawsneurallanguage, narayanan2021efficient}, which focuses on per-token inference cost dominated by the forward pass through the transformer stack. The primary components include self-attention and feed-forward networks (FFNs).

The relevant architectural configurations of the two models are summarized as follows:

\begin{table}[ht]
\centering
\resizebox{\linewidth}{!}{%
\begin{tabular}{lcc}
\toprule
 & \textbf{DeepSeek-R1-Distill-Qwen-1.5B} & \textbf{QwQ-32B} \\
\midrule
Hidden size $d$           & 1536  & 5120 \\
Number of layers $L$      & 28    & 64 \\
Attention heads           & 12    & 40 \\
FFN hidden size $d_{\text{ffn}}$ & 8960  & 27648 \\
\bottomrule
\end{tabular}%
}
\caption{Model architectural parameters.}
\end{table}

We adopt a standard transformer-based decoder-only FLOPs estimation, considering cached key/value states. Thus, the FLOPs per token (forward pass only) is approximated as:

\[
\text{FLOPs}_{\text{token}} \approx L \cdot \left( 4d^2 + 8d \cdot d_{\text{ffn}} + 2d \cdot T \right)
\]

where $L$ is the number of layers, $d$ the hidden size, $d_{\text{ffn}}$ the intermediate feedforward size, and $T$ the input sequence length. For autoregressive generation with KV cache, we assume both models receive the same context length $T$, and we fix the generation length within 8192 tokens in our experimental setting. We thus estimate the FLOPs ratio ($\alpha$) per token as:
\[
\alpha = \frac{F_{d}}{F_{t}} \approx 0.042 \sim 0.048
\]

\noindent indicating that QwQ-32B requires approximately $21 \sim 24 \times$ more FLOPs per token than DeepSeek-R1-Distill-Qwen-1.5B.

\noindent \textbf{Note.} This analysis assumes standard attention and FFN computation without low-rank optimizations, quantization, or sparsity. Decoder-only inference with cached KV is considered. While wall-clock latency can differ based on implementation (e.g., FlashAttention, tensor parallelism), the FLOPs metric remains a robust theoretical comparison for compute cost.

\subsection{Normalized FLOPs}

\noindent\textbf{(1) Baseline Inference.}
A single invocation of the target model generates $T_{\mathrm{base}}$ tokens, consuming

\begin{equation}
\mathrm{FLOPs}_{\mathrm{base}}
\;=\;
T_{\mathrm{base}}\;F_{t}.
\end{equation}

By definition, we normalize this quantity to unity:

\begin{equation}
\boxed{
\mathrm{\gamma}_{\mathrm{base}}
\;=\;
1.
}
\end{equation}

\noindent\textbf{(2) Traditional Parallel Inference.}
For parallel inference across $N$ independent paths, each producing $T_{\mathrm{base}}$ tokens, the total compute is:

\begin{equation}
\mathrm{FLOPs}_{\mathrm{parallel}} = N T_{\mathrm{base}} F_{t}
\end{equation}
\begin{equation}
\Longrightarrow \boxed{\mathrm{\gamma}_{\mathrm{parallel}} = N}
\end{equation}

\noindent\textbf{(3) Speculative Parallel Inference.}
To estimate the compute cost, we approximate the token-level rewrite rate by the step-level rewrite rate. For each of the $N$ inference paths, the draft model first generates $T$ tokens at a cost of ($T\,F_{d}$). The target model then selectively rewrites a fraction $R$ of these tokens, incurring an additional cost of ($R\,T\,F_{t}$). Tokens that are only scored but not rewritten contribute negligible compute and are thus ignored in our analysis. The total per-path cost is therefore:

\begin{equation}
T\,F_{d}
\;+\;
R\,T\,F_{t}
\;=\;
T\,F_{t}\,\bigl(\alpha + R\,\bigr).
\end{equation}

Aggregated over $N$ paths and normalized by the baseline cost, we obtain

\begin{equation}
\begin{split}
\mathrm{\gamma}_{\mathrm{spec}}
&= \frac{N\,\bigl(T\,F_{d} + R\,T\,F_{t}\bigr)}{T_{\mathrm{base}}\,F_{t}} \\
&= N\,\beta\,\bigl(R + (1-R)\,\alpha\bigr),
\end{split}
\end{equation}

\noindent where we have used $\beta=T/T_{\mathrm{base}}$ and $\alpha=F_{d}/F_{t}$.  In compact form:

\begin{equation}
\boxed{
\mathrm{\gamma}_{\mathrm{spec}}
=
N\,\beta\,\bigl(R + (1-R)\,\alpha\bigr).
}
\end{equation}

\noindent These three closed‐form expressions allow direct, quantitative comparison of the relative computational burden of baseline, traditional parallel, and speculative parallel inference—independently of hardware or software idiosyncrasies.

\section{Threshold Sensitivity}

To better understand the effect of acceptance threshold in our SSR, we conduct an ablation study on the AIME dataset by varying the semantic acceptance threshold from low to high values. For each setting, we measure the final reasoning accuracy (pass@1) and the inverse of normalized Flops ($1/\gamma$), which serves as an indicator of inference efficiency.

\begin{figure}[ht]
  \centering
  \includegraphics[width=0.9\linewidth]{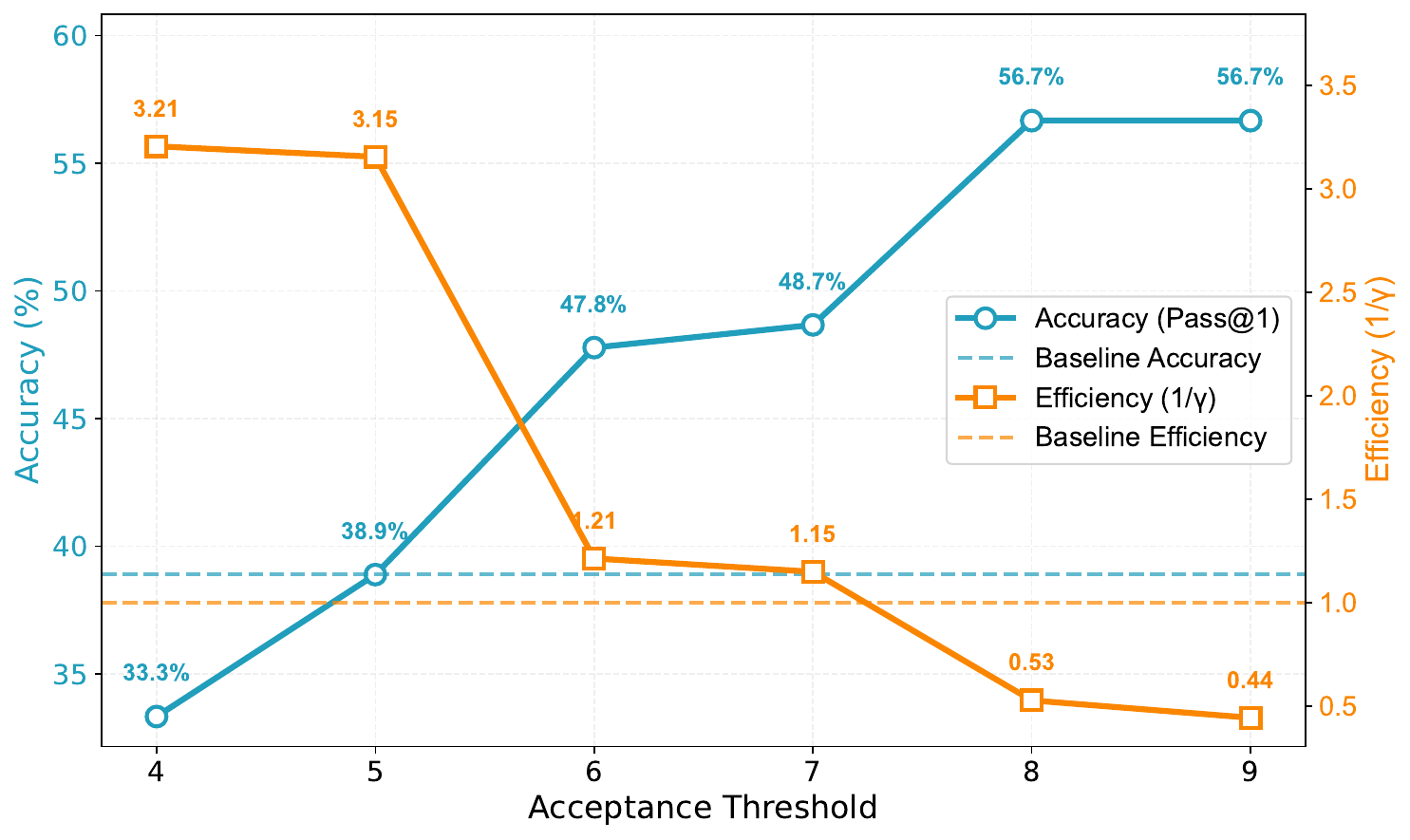}
  \caption{Effect of acceptance threshold on reasoning accuracy and efficiency. Evaluated on the AIME dataset using SSR with 3 parallel reasoning paths. Higher thresholds yield better accuracy but incur significantly more compute. Threshold = 7 achieves the best trade-off.}
  \label{fig:threshold_ablation}
\end{figure}

As shown in Figure~\ref{fig:threshold_ablation}, increasing the threshold consistently improves the correctness of accepted reasoning steps, as measured by pass@1 accuracy. However, this also leads to a rapid degradation in efficiency, since more speculative steps are rejected and require regeneration by the target model.

To capture the trade-off, we plot both accuracy and inverse normalized FLOPs. We find that threshold = 7 offers a sweet spot, providing significant gains in accuracy while still maintaining compute cost below the baseline. This threshold also aligns well with the scoring distribution analysis shown in the main paper (see Section Experiments), reinforcing our selection.

\section{Effect of Strategy Pool Size}

\begin{figure}[ht]
  \centering
  \includegraphics[width=0.9\linewidth]{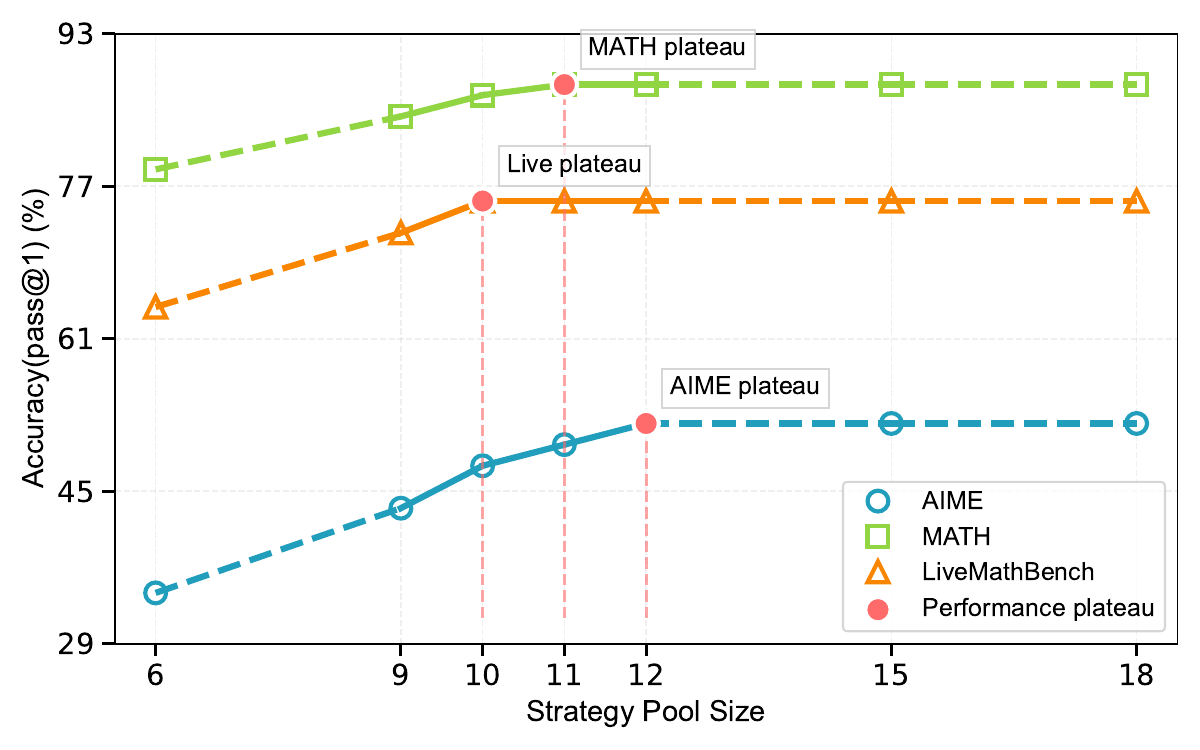}
  \caption{Effect of strategy pool size on pass@1 accuracy. Evaluated on the AIME dataset with 3 parallel paths.}
  \label{fig:strategy_pool_size}
\end{figure}

We evaluate the impact of the strategy pool size in SSR by running experiments on the AIME dataset, fixing the number of parallel reasoning paths to 3 and varying the pool size from 6 to 18. Accuracy is measured via pass@1, and results are plotted in Figure~\ref{fig:strategy_pool_size}.

As shown in the figure, accuracy improves steadily as the pool size increases up to 12, but further enlarging the pool yields negligible gains. This indicates that increasing strategy diversity beyond this level is largely unnecessary for math reasoning tasks such as AIME. However, for broader domains involving more diverse reasoning patterns, a larger strategy pool may be beneficial to capture task-specific reasoning styles.

\section{Prompt Templates and Strategy Pool}
\label{appendix:prompt}
To enable robust and diverse mathematical reasoning, we construct a strategy pool comprising distinct problem-solving approaches tailored to algebra, geometry, number theory, and combinatorics. Each strategy corresponds to a specific reasoning paradigm and is mapped to a dedicated prompt that guides the model's thought process. The pool is designed for broad coverage of problem types and remains extensible for future integration of new strategies.

The \textsc{SPM} framework executes multiple strategies in parallel, treating each as an independent reasoning path. This design improves solution diversity, enhances robustness to problem variations, and increases the likelihood of identifying correct or insightful answers.

\noindent\textbf{Strategy A. Algebraic simplification}
\begin{quote}
\texttt{Use algebraic manipulation (expansion, factoring, substitution) to simplify the expressions or equations.}
\end{quote}

\noindent\textbf{Strategy B. Clever substitution}
\begin{quote}
\texttt{Use a smart change of variables to transform the problem into a simpler or standard form.}
\end{quote}

\noindent\textbf{Strategy C. Coordinate geometry}
\begin{quote}
\texttt{Introduce a coordinate system and use analytic geometry techniques (e.g. distance, slope, midpoint).}
\end{quote}

\noindent\textbf{Strategy D. Complex numbers in geometry}
\begin{quote}
\texttt{Use complex number representation for points to solve geometric problems.}
\end{quote}

\noindent\textbf{Strategy E. Number theory}
\begin{quote}
\texttt{Apply modular arithmetic, divisibility, prime factorization, or Diophantine techniques.}
\end{quote}

\noindent\textbf{Strategy F. Combinatorics}
\begin{quote}
\texttt{Count the number of arrangements, selections, or outcomes using combinatorial principles.  
Strategy K. Casework or constructive examples: Systematically enumerate or construct possible cases to exhaust the possibilities.}
\end{quote}

\noindent\textbf{Strategy G. Probability}
\begin{quote}
\texttt{Use probability models, expected value, or case enumeration to compute probabilities.}
\end{quote}

\noindent\textbf{Strategy H. Functional equations}
\begin{quote}
\texttt{Analyze and solve equations involving functions and their values under certain operations.}
\end{quote}

\noindent\textbf{Strategy I. Recursion or invariants examples}
\begin{quote}
\texttt{Identify recursive patterns or quantities that remain invariant under operations.}
\end{quote}

\noindent\textbf{Strategy J. Geometry}
\begin{quote}
\texttt{Use classical Euclidean geometry (angles, lengths, similarity, etc.) and synthetic arguments.}
\end{quote}

\noindent\textbf{Strategy K. Casework or constructive examples}
\begin{quote}
\texttt{Systematically enumerate or construct possible cases to exhaust the possibilities.}
\end{quote}

\noindent\textbf{Strategy L. Calculus or inequalities}
\begin{quote}
\texttt{Use derivatives, bounds, or inequality techniques like AM-GM or Cauchy-Schwarz.}
\end{quote}

\noindent\textbf{Strategy M. Unknow}
\begin{quote}
\texttt{I cannot confidently determine which strategy is suitable from the list above.}
\end{quote}

\end{document}